\documentclass[runningheads]{llncs}

\usepackage{eccv}

\usepackage{eccvabbrv}

\usepackage{graphicx}
\usepackage{booktabs}
\usepackage{xcolor}
\usepackage{colortbl}
\usepackage{adjustbox}
\usepackage{xcolor}

\usepackage{amsmath, amssymb, amsfonts}
\usepackage{booktabs}   % \toprule \midrule \bottomrule \cmidrule
\usepackage{multirow}   % \multiro
\usepackage{pifont}   % 提供 \ding{55}
\usepackage{bbding}

\usepackage{hyperref}
\usepackage{makecell}
\usepackage{enumitem}

\usepackage{subcaption} % provides subfigure environment
\usepackage[accsupp]{axessibility}  % Improves PDF readability for those with disabilities.

\usepackage{orcidlink}

\begin{document}

\title{DGSeg: Dynamic Gating of Semantic--Spatial Guided Predictions for Reasoning Segmentation}

\titlerunning{DGSeg}

\author{Ruizhe Zeng\inst{1,2}\orcidlink{0009-0004-3513-2297} \and
  Siyu Cao\inst{1,2}\orcidlink{0009-0002-3024-4500} \and
Lu Zhang\inst{1,2}\textsuperscript{*}\orcidlink{0000-0001-6240-5300} \and
Zhi-yong Liu\inst{1,2,3}\textsuperscript{*}\orcidlink{0000-0003-2148-1846}}

\authorrunning{Zeng et al.}

\institute{State Key Laboratory of Multimodal Artificial Intelligence Systems,\\
Institute of Automation, Chinese Academy of Sciences,\\
Beijing 100190, China\\
\email{\{zengruizhe2022, caosiyu2024, lu.zhang, zhiyong.liu\}@ia.ac.cn}\and
School of Artificial Intelligence, University of Chinese Academy of Sciences,\\
Beijing 100049, China \and
Nanjing Artificial Intelligence Research of IA, Nanjing, 211100, China}

\maketitle

\begin{abstract}
Reasoning segmentation aims to predict pixel-wise masks for targets given complex language queries. Existing approaches leverage Multimodal Large Language Models (MLLMs) for vision-language reasoning and generate intermediate target cues (e.g., points or boxes) to guide a segmentation model. However, compressing rich reasoning into sparse cues often introduces ambiguity and noise, preventing these cues from accurately preserving the reasoning intent. While multiple complementary cues can enrich target information, existing methods typically feed them jointly into a single segmentation process, allowing ambiguous or erroneous cues to affect the entire prediction. Therefore, we propose \textbf{DGSeg}, a reasoning segmentation framework that learns to fuse predictions guided by semantic and spatial cues. Specifically, the MLLM jointly reasons about  both target identity and spatial location, producing complementary semantic and spatial cues that are fed into separate segmentation branches.  Their predictions are adaptively integrated by a lightweight dynamic gating module trained with relative branch-quality supervision to suppress noisy or conflicting regions. Extensive experiments demonstrate that DGSeg consistently outperforms strong baselines on multiple benchmarks and achieves 69.6\% and 67.3\% gIoU on the challenging ReasonSeg validation and test splits. Code is available at \href{https://github.com/RZZeng/DGSeg}{https://github.com/RZZeng/DGSeg}.
\end{abstract}

\section{Introduction}
\label{sec:intro}
Reasoning segmentation is the task of segmenting objects in an image given an implicit language query~\cite{lai2024lisa}. For instance, for ``find a tool that can tighten screws'', the model must reason about the functionality of entities in the scene, identify a suitable object (e.g., a screwdriver), and precisely segment it. Therefore, this task is crucial for vision systems that need to translate high-level human intent into object regions, especially in embodied settings where downstream interaction or manipulation depends on precise localization~\cite{vuong2024language, collins2024forcesight}.

\begin{figure}[t]
  \centering
  \includegraphics[width=\linewidth]{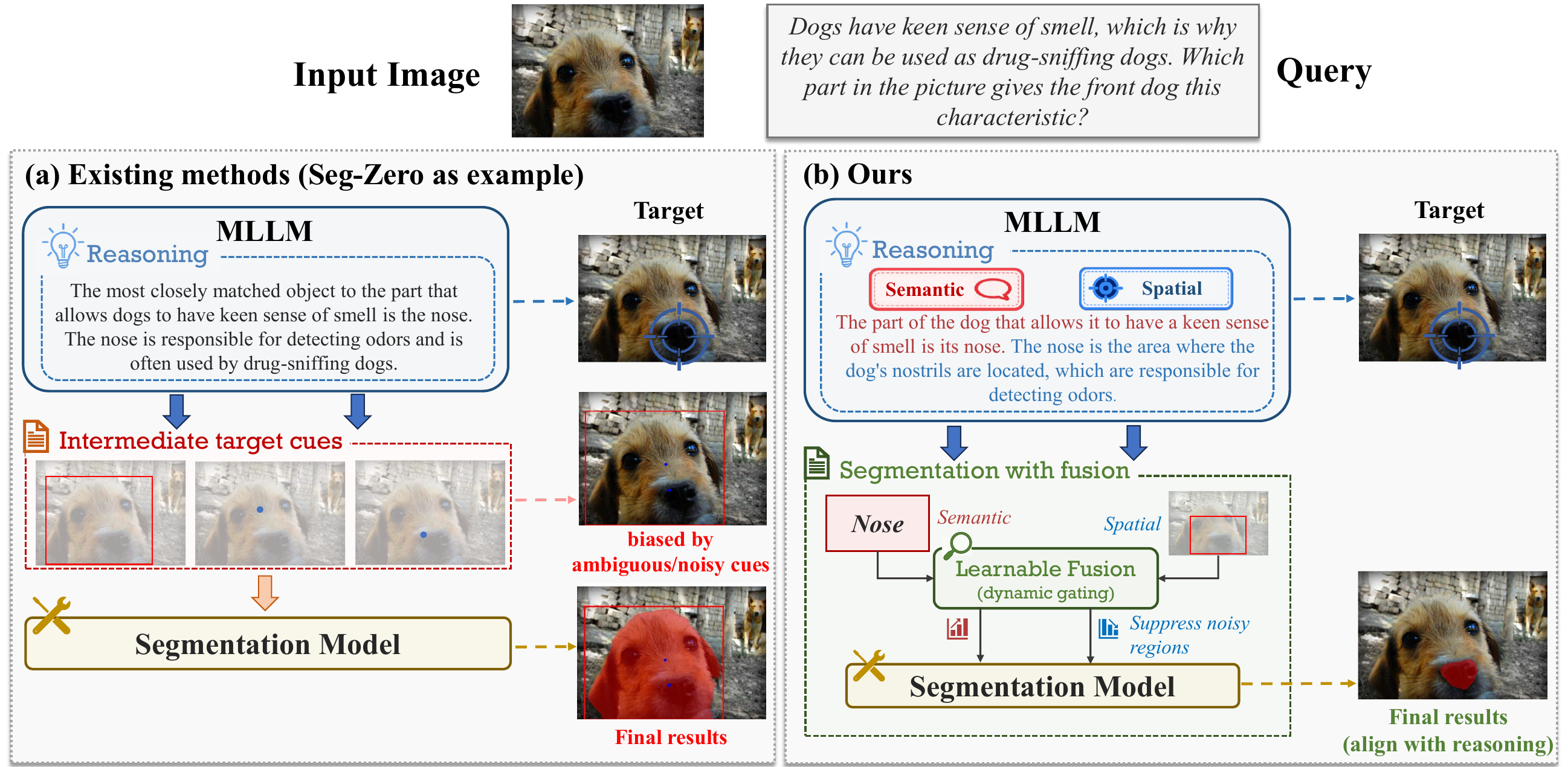}
  \caption{\textbf{Motivation of our method.} Existing approaches (e.g., Seg-Zero~\cite{liu2025seg}) use an MLLM to generate target cues for downstream segmentation. However, they feed cues directly without mitigating potential noise, causing results to deviate from reasoning due to ambiguous or noisy cues (e.g., oversized boxes or erroneous points). In contrast, we leverage complementary semantic--spatial reasoning and a learnable fusion process to resolve potential noise, helping the final results better align with the reasoning intent.}
  \label{fig1}
\end{figure}

While Multimodal Large Language Models (MLLMs)~\cite{bai2023qwen, bai2025qwen2, liu2023visual, chen2024internvl} exhibit remarkable reasoning capabilities, they struggle with fine-grained visual localization tasks~\cite{he2025analyzing, yao2025argus, wang2025cof}. To leverage their ability to analyze complex queries while addressing this limitation, current methods typically combine MLLMs with promptable segmentation models, and bridge high-level MLLM reasoning and the prompt space of the segmentation model by generating intermediate target cues (e.g., latent token embeddings~\cite{lai2024lisa, bao2024cores, qian2025reasoning}, points, or bounding boxes~\cite{ravi2024sam, liu2025seg, you2025seg}). However, compressing rich reasoning into such target cues may degrade referential details and make the cues ambiguous or noisy, preventing them from accurately preserving the reasoning intent.  As shown in \cref{fig1}, when the target is a dog's nose, the bounding box generated by the MLLM may loosely cover the nose while also including irrelevant regions (e.g., the entire dog), and the segmentation model may take these regions as valid targets.  Although multiple complementary cues can enrich target information, current methods typically encode them into a unified representation for segmentation, where any ambiguous or erroneous cue may introduce noise into the whole segmentation process.  While some recent works~\cite{dusam, he2025rsagent} have attempted to iteratively refine these cues using MLLMs, they often require prohibitive costs in both training and inference. Thus, developing an efficient approach to reduce target referential bias caused by ambiguous or noisy cues remains a critical challenge.

To address these challenges, we propose a novel reasoning segmentation pipeline that builds learnable adaptive fusion upon separate processing of complementary target cues. Unlike existing methods that directly encode all cues into a unified segmentation process, our core insight is to isolate potential noise by independently processing complementary cues, and to guide adaptive fusion with supervision derived from the relative segmentation quality of different branches. In this way, the pipeline integrates effective information from different cues while mitigating referential bias and preserving the original reasoning intent.

Based on this pipeline, we instantiate our framework as \textbf{DGSeg}, where dynamic gating serves as a lightweight implementation of the learnable fusion mechanism. The classic view posits that visual understanding amounts to knowing \textit{what is where}~\cite{man1982computational}, suggesting that object representation requires both semantic identity and spatial localization. Following this insight, DGSeg guides the MLLM to jointly reason about \emph{what} the target is and \emph{where} it is, producing an explicit textual description and a bounding-box level localization as complementary semantic and spatial cues. These cues are processed by separate semantic and spatial segmentation branches, respectively. Finally, the dynamic gating module aggregates their predictions using pixel-wise fusion weights estimated from branch features.

Extensive experiments on referring and reasoning segmentation benchmarks demonstrate that DGSeg consistently surpasses a wide range of strong baselines. Under the zero-shot setting, it achieves 66.0\% and 60.0\% gIoU with the 3B-scale model and 69.6\% and 67.3\% gIoU with the 7B-scale model on the challenging ReasonSeg validation and test splits~\cite{lai2024lisa}. Ablation studies further confirm the effectiveness of the proposed pipeline, showing that both semantic-spatial cue generation and learned fusion bring consistent performance gains.   Moreover, fusion behavior analysis indicates that the learned fusion module mitigates relative noise across dual branches by adaptively assigning higher weights to more accurate predictions, making the final segmentation results more consistent with the MLLM’s intended reasoning.

Overall, our main contributions are summarized as follows:
\begin{itemize}[topsep=2pt,itemsep=2pt,parsep=0pt]
    \item We propose a new reasoning segmentation pipeline that learns to adaptively fuse predictions guided by complementary target cues, thereby reducing the impact of ambiguous or erroneous cues and better preserving the original reasoning intent.
    \item We instantiate this pipeline as DGSeg, which generates semantic and spatial cues from the MLLM, processes them with separate segmentation branches, and aggregates their predictions using a lightweight dynamic gating module.
    \item Extensive experiments across referring and reasoning segmentation benchmarks show consistent gains over strong baselines, achieving up to 69.6\% and 67.3\% gIoU on the challenging ReasonSeg validation and test splits. Ablation studies and further analyses  validate the effectiveness of the proposed pipeline and DGSeg.
\end{itemize}

\section{Related Works}

\subsection{Reasoning Segmentation}
Reasoning segmentation is the task that focuses on extracting the implicit referential intent in a query and grounding it into a pixel-level segmentation mask. Existing methods typically leverage an MLLM to perform vision-language reasoning and then pass the inferred target cues to a downstream segmentation model. These methods can be broadly categorized into three categories.

The first category uses embeddings of special tokens to prompt the downstream segmentation model. LISA~\cite{lai2024lisa} introduces a special \texttt{<SEG>} token and uses its contextual embedding from the MLLM reasoning process to condition SAM for mask prediction. CoReS~\cite{bao2024cores} follows a dual-chain pipeline that first coarsely localizes the target region and then refines it into a fine-grained segmentation mask. The second line exploits intermediate signals produced during MLLM reasoning (e.g., vision-language attention maps) to extract referential information. For example, Kang et al.~\cite{kang2025your} propose a training-free reasoning segmentation method by devising hand-crafted rules to extract localization cues from attention maps. LENS~\cite{liu2025segmentation} further introduces a plug-and-play framework that attaches a lightweight segmentation head to refine spatial information in attention maps. The third line leverages the direct textual output of MLLM reasoning as prompts to condition the downstream segmentation model. These methods typically finetune the MLLM via supervised finetuning (SFT)~\cite{chen2024sam4mllm} or reinforcement learning (RL)~\cite{liu2025seg, liu2025visionreasoner, zhou2026reasoning, you2025seg} based on the quality of generated textual outputs, while keeping the downstream segmentation model frozen.

Despite their differences, these methods often rely on a single type of target cue and directly use it to condition mask prediction. As a result, the referent can remain ambiguous or affected by noisy cues, leading to ambiguity and error propagation in the final masks.

\subsection{Multimodal Large Language Model}

Early research on vision-language foundation models~\cite{radford2021learning, jia2021scaling, li2021align} primarily focused on aligning image features with language for a wide range of downstream perception tasks. Recently, extending the language understanding and reasoning capabilities of LLMs~\cite{touvron2023llama, bai2023qwen1} to multimodal inputs has offered a new paradigm for tackling vision-language problems. Representative works like LLaVA~\cite{liu2023visual} show that treating visual features as tokens and finetuning on high-quality multimodal instruction data can effectively elicit generalizable vision-language capabilities. Recent open-source MLLMs~\cite{dai2023instructblip, wang2024qwen2, bai2025qwen2, chen2024expanding, chen2024spatialvlm} further strengthen recognition, grounding, document understanding, and long-context multimodal reasoning through improved pretraining data, post-training recipes, and scaling strategies. For fine-grained output tasks such as segmentation, existing methods~\cite{lai2024lisa, hu2024visual, yan2024visa} commonly couple MLLMs with specialized perception models to leverage the strong reasoning capabilities of MLLMs. In this paradigm, a key challenge is to obtain target cues that are both complete and reliable for guiding the downstream segmentation model, which directly motivates our method.

\subsection{Promptable Segmentation}

Promptable segmentation predicts segmentation masks conditioned on flexible user prompts. A representative work in this line is SAM~\cite{kirillov2023segment}, which supports sparse geometric prompts such as points and bounding boxes, and dense prompts in the form of coarse masks. It encodes the image and prompts separately and then produces the final segmentation via a mask decoder. Trained on the large-scale SA-1B dataset~\cite{kirillov2023segment}, SAM demonstrates strong zero-shot segmentation capability and excellent fine-grained visual perception. Subsequent work, SAM2~\cite{ravi2024sam}, extends promptable segmentation to video object tracking, while SAM3~\cite{carion2025sam} introduces a DETR-based detector~\cite{carion2020end} to detect and segment all visual elements referred to by textual queries or visual exemplars.

Despite their strong generalization, promptable segmentation models are sensitive to the quality of input prompts. Stable-SAM~\cite{fan2023stable} reveals that the mask decoder tends to produce biased feature activations given low-quality prompts and presents a novel deformable sampling plugin to improve prompt stability. SAMRefiner~\cite{lin2025samrefiner} introduces a prompt excavation strategy to mine diverse and accurate prompts from coarse masks for segmentation refinement. However, most existing reasoning segmentation approaches overlook the adverse impact of prompt quality, making their predictions susceptible to noise. Although iterative MLLM refinement has been explored to reduce noisy prompts~\cite{dusam, zhu2025segagent}, it is computationally expensive and remains vulnerable to hallucinated cues. We therefore adopt a learnable fusion module to evaluate pixel-level features from the segmentation process and integrate dual-branch predictions.

\section{Method}

In this section, we present DGSeg as a concrete implementation of the proposed pipeline, which independently processes complementary cues and learns to adaptively fuse their predictions.  We first describe how the MLLM generates semantic and spatial cues. We then introduce the dual-branch segmentation design and the dynamic gating module for adaptive prediction fusion. Finally, we detail the two-stage optimization strategy for cue generation and fusion learning.

\subsection{Architecture of DGSeg}
\begin{figure}[t]
  \centering
  \includegraphics[width=\linewidth]{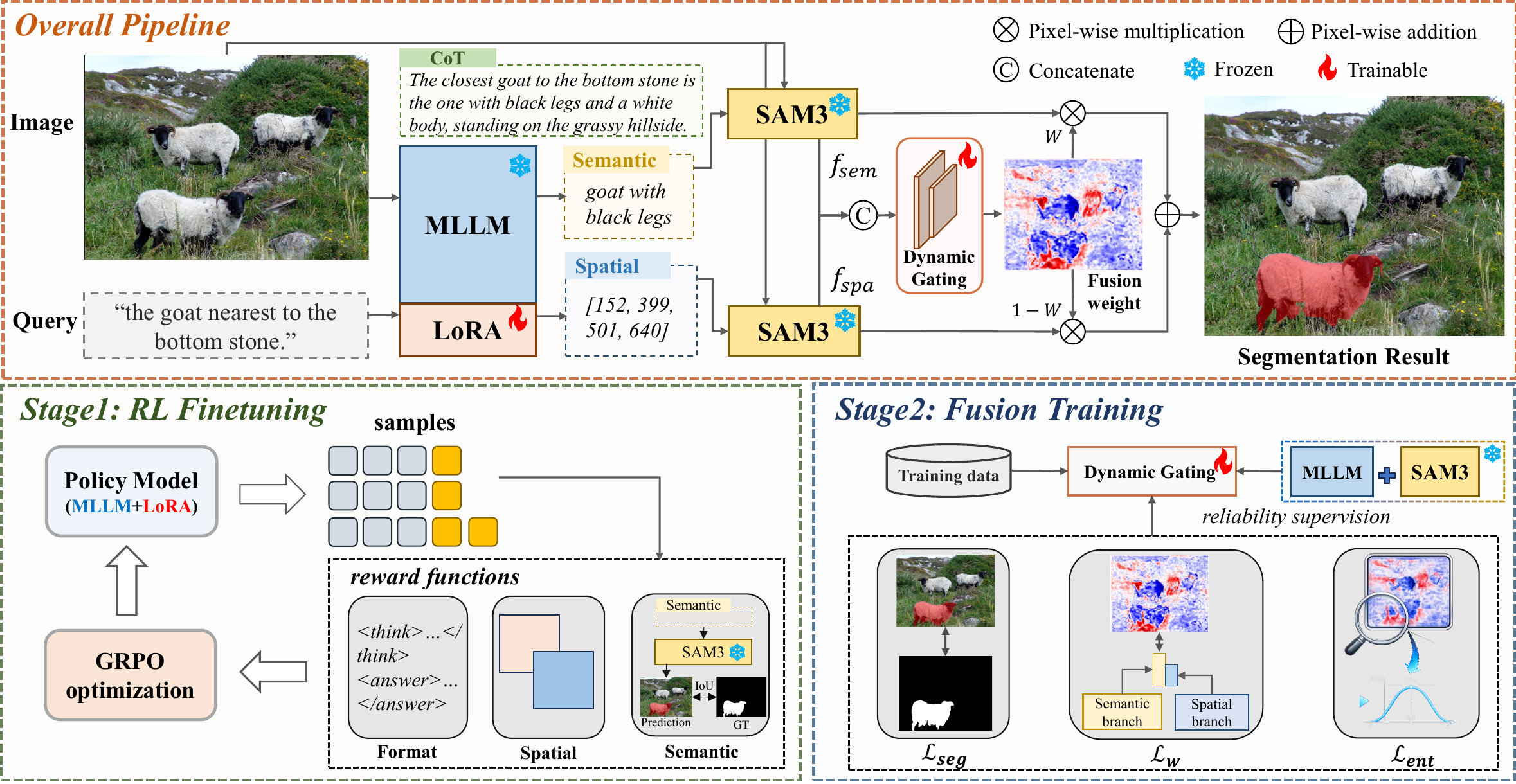}
  \caption{\textbf{Overall framework of DGSeg.} Given an image and a language query, the MLLM first generates complementary semantic and spatial cues. These cues are processed by separate segmentation branches to produce initial results. A learnable dynamic gating module evaluates the features from dual-branch segmentation and adaptively fuses these results.}
  \label{fig2}
\end{figure}

\textbf{Framework Overview.} Our framework is illustrated in \cref{fig2}, which consists of a reasoning model as the target cue generator and a downstream segmentation model as the cue executor. Given an input RGB image $I \in \mathbb{R}^{H \times W \times 3}$ and a language query $T$, the reasoning model performs high-level reasoning to produce semantic--spatial cues. The segmentation model then processes these cues through two branches and employs a dynamic gating module to fuse predictions.

\noindent\textbf{Reasoning Model.} Given the image–query pair $\{I, T\}$, we instruct the reasoning model $F_{\text{reason}}$ to infer the intended target from two complementary aspects: semantic identity and spatial localization. Inspired by chain-of-thought prompting~\cite{wei2022chain}, which encourages models to articulate intermediate reasoning steps before producing final outputs, we guide $F_{\text{reason}}$ to first generate an explicit reasoning trace $c_{\text{CoT}}$. The model then outputs a concise textual description $c_{\text{sem}}$ and a bounding-box-level localization $c_{\text{spa}}$ for the intended target, i.e.,
\begin{equation}
  c_{\text{CoT}}, c_{\text{sem}}, c_{\text{spa}} = F_{\text{reason}}(I, T).
\end{equation}
Compared with prior approaches that rely solely on geometric prompts such as points,  bounding boxes~\cite{liu2025seg, you2025seg, huang2025sam} or latent embeddings of special tokens~\cite{lai2024lisa, yuan2025sa2va}, our formulation characterizes the target from both semantic and spatial perspectives and provides a more complete target representation for downstream segmentation.

\noindent\textbf{Segmentation Model.} We adopt SAM3~\cite{carion2025sam} as the segmentation model $F_{\text{seg}}$ since it supports both semantic and spatial inputs. In the original SAM3 framework, multiple prompts are encoded jointly into a single unified prompt representation, which allows unreliable prompts from any source to affect the entire prediction. To mitigate this issue, DGSeg processes the semantic cue $c_{\text{sem}}$ and spatial cue $c_{\text{spa}}$ via separate branches to produce mask logits $\ell_{\text{sem}}, \ell_{\text{spa}} \in \mathbb{R}^{H \times W}$ and the corresponding mask predictions $m_{\text{sem}}$ and $m_{\text{spa}}$.

\noindent\textbf{Dynamic Gating Module.}
After the semantic and spatial branches produce their predictions, we introduce a lightweight dynamic gating module $F_{\text{dg}}$ to adaptively estimate pixel-wise fusion weights for fusing the dual-branch outputs. The input to $F_{\text{dg}}$ is the feature embedding $f \in \mathbb{R}^{1\times c\times h\times w}$ from the pixel decoder of SAM3~\cite{carion2025sam}, which encodes the joint representation of the image and the cues and preserves informative segmentation patterns such as boundary structures and high-response regions.
Then, we concatenate the pixel embeddings $f_{\text{sem}}$ and $f_{\text{spa}}$ from the semantic and spatial branches, and feed them into a lightweight network composed of two convolutional layers to jointly evaluate the features from the two branches and predict a pixel-wise fusion weight map $W$:
\begin{equation}
  W = \sigma(F_{\text{dg}} (\text{Concate}(f_{\text{sem}}, f_{\text{spa}}))) \in [0,1]^{1\times h\times w},
\end{equation}
where $\sigma$ denotes the sigmoid activation function, and $h$ and $w$ denote the spatial resolution of the feature embeddings.

We then resize $W$ to match the spatial resolution of the logit maps (denoted as $W^{\uparrow}$), and compute the final logit and mask prediction $\{\ell, m\}$ by adaptively combining the logits from the two branches:
\begin{equation}
  \ell = W^{\uparrow} \odot \ell_{\text{sem}} + (1-W^{\uparrow})\odot \ell_{\text{spa}},
\end{equation}
\begin{equation}
  m = \mathbb{I}\!\left[\ell > 0\right],
\end{equation}
where $\odot$ denotes element-wise multiplication and $\mathbb{I}[\cdot]$ is the indicator function.

\subsection{Training Strategy}
\textbf{Strategy Overview.} We introduce a two-stage training strategy to optimize DGSeg effectively across both cue generation and mask prediction. In the first stage, we freeze the segmentation model and finetune the MLLM via reinforcement learning to produce accurate semantic and spatial cues. In the second stage, we freeze the MLLM and the segmentation backbone to train the dynamic gating module specifically through supervised learning. This separated design ensures stable supervision for adaptively fusing dual-branch predictions and prevents interference between optimizing the cue generator and the segmentation executor.

\noindent\textbf{Stage 1: RL Finetuning.} In this stage, we finetune the MLLM with GRPO~\cite{shao2024deepseekmath} to improve target cue generation while preserving generalization. Our objective is to ensure the reasoning outputs are compatible with the segmentation model and reliable for identifying the intended target. Specifically, the MLLM is encouraged to produce structured outputs that can be parsed into extractable cues which can accurately refer to the target. To this end, we optimize a composite reward comprising three terms, including a format reward for valid structured outputs and semantic and spatial rewards that assess cue reliability. We detail each reward component below.

\begin{itemize}
  \item \textbf{Format reward.} We require the MLLM to place its explicit reasoning process within \mbox{\texttt{\detokenize{<think></think>}}} tags and the final response within \texttt{<answer>}\allowbreak\texttt{</answer>} tags so that the target cues can be reliably parsed. Let the MLLM output be $y$,  we define
    \begin{equation}
      \mathcal{R}_{\text{format}}=
      \begin{cases}
        1, & \text{if } y \text{ follows the predefined output format},\\
        0, & \text{otherwise}.
      \end{cases}
    \end{equation}

  \item \textbf{Spatial reward.} To encourage accurate coarse localization, we use bounding-box IoU as the reward. Given the predicted bounding box target cue \(c_{\text{spa}}\) and the ground-truth box \(b^{\ast}\), we define
    \begin{equation}
      \mathcal{R}_{\text{spatial}}
      =\mathrm{IoU}(c_{\text{spa}},b^{\ast}) = \frac{|c_{\text{spa}} \cap b^{\ast}|}{|c_{\text{spa}} \cup b^{\ast}|}.
    \end{equation}

  \item \textbf{Semantic reward.} In image classification, semantic correctness is often evaluated by matching a predicted category to a ground truth label~\cite{deng2009imagenet}. However, category labels are costly to annotate and may not be well-aligned with the semantic prompt space that the segmentation model operates on. Inspired by human feedback alignment~\cite{christiano2017deep}, we define a semantic reward based on the mask preference of the segmentation model. Specifically, we use the extracted semantic cue  $c_{\text{sem}}$ to obtain the prediction $m_{\text{sem}}$  from the semantic branch. The semantic reward is then defined as the IoU between $m_{\text{sem}}$ and the ground truth $m^{\ast}$:
    \begin{equation}
      \mathcal{R}_{\text{semantic}}= \text{IoU}(m_{\text{sem}}, m^{\ast}) = \frac{|m_{\text{sem}} \cap m^{\ast}|}{|m_{\text{sem}} \cup m^{\ast}|}.
    \end{equation}
\end{itemize}

\noindent\textbf{Stage 2: Fusion Training.} In this stage, we freeze the MLLM and the segmentation backbone to train only the dynamic gating module $F_{\text{dg}}$ for optimal prediction fusion. Since the final segmentation outcome is jointly determined by cue quality and gating weights, optimizing segmentation loss $\mathcal{L}_{\text{seg}}$ alone provides ambiguous supervision for dual-branch fusion. We therefore add an auxiliary target that encourages the gate to favor the branch that is closer to the ground truth. Specifically, given the two branch masks $m_{\text{sem}}$ and $m_{\text{spa}}$ and the ground-truth mask $m^{\ast}$, we compute
\begin{equation}
  s_{\text{sem}} = \text{IoU}(m_{\text{sem}}, m^{\ast}), \qquad
  s_{\text{spa}} = \text{IoU}(m_{\text{spa}}, m^{\ast}),
\end{equation}
and derive a soft fusion target weight as
\begin{equation}
  W^{\ast}
  =\left(\frac{\exp\!\left(s_{\text{sem}}/\tau\right)}
  {\exp\!\left(s_{\text{sem}}/\tau\right)+\exp\!\left(s_{\text{spa}}/\tau\right)}\right)^{H\times W},
\end{equation}
where $\tau$ controls the sharpness. We take $W^{\ast}$ to supervise the predicted weights $W^{\uparrow}$ with a binary cross-entropy loss $\mathcal{L}_{\text{w}}$, and an entropy penalty $\mathcal{L}_{\text{ent}}$ to discourage overly averaged weights:
\begin{equation}
  \mathcal{L}_{\text{ent}} =
  - W^{\uparrow}\log(W^{\uparrow}) - (1-W^{\uparrow})\log(1-W^{\uparrow}).
\end{equation}
Finally, the overall training objective is
\begin{equation}
  \mathcal{L}
  = \mathcal{L}_{\text{seg}}
  + \lambda_{\text{w}}\,\mathcal{L}_{\text{w}}
  + \lambda_{\text{ent}}\,\mathcal{L}_{\text{ent}}.
\end{equation}
During training, we anneal $\lambda_{\text{w}}$ over iterations $t$ to provide strong guidance early while avoiding convergence to the soft target. Finally, we emphasize pixels where fusion matters most by reweighting the loss, assigning higher weights to pixels where the two branches disagree compared to those with consistent predictions. Detailed configurations for these weight hyperparameters are provided in the supplementary material.

\begin{table}[t]
  \centering
  \small
  \caption{Performance comparison on ReasonSeg benchmark under the \textbf{zero-shot setting}.}
  \setlength{\tabcolsep}{2.7pt}
  \renewcommand{\arraystretch}{1.05}

  \makebox[0.85\linewidth][c]{%
    \begin{tabular}{p{3.35cm} >{\centering\arraybackslash}p{2.85cm} c c c c}
      \toprule
      \multirow{2}{*}{\textbf{Method}} &
      \multirow{2}{*}{\textbf{MLLM}} &
      \multicolumn{2}{c}{\textbf{Val}} &
      \multicolumn{2}{c}{\textbf{Test}} \\
      \cmidrule(lr){3-4}\cmidrule(lr){5-6}
      & & gIoU & cIoU & gIoU & cIoU \\
      \midrule
      OVSeg~\cite{liang2023open}          & --             & 28.5 & 18.6 & 26.1 & 20.8 \\
      SEEM~\cite{zou2023segment}          & --             & 25.5 & 21.2 & 24.3 & 18.7 \\
      Grounded-SAM~\cite{ren2024grounded} & --             & 26.0 & 14.5 & 21.3 & 16.4 \\
      \midrule
      Seg-Zero~\cite{liu2025seg}          & Qwen2.5-VL-3B   & \underline{62.6} & \underline{58.5} & 56.1 & 48.6 \\
      Seg-R1~\cite{you2025seg}            & Qwen2.5-VL-3B   & 60.8 & 56.2 & 55.3 & 46.6 \\
      RISE~\cite{zhou2026reasoning}               & Qwen2.5-VL-3B   & 52.6 & 43.1 & 52.9 & 45.1 \\
      PIXELTHINK~\cite{wang2025pixelthink}& Qwen2.5-VL-3B   & 62.3 & \underline{58.5} & \underline{58.8} & \underline{52.1} \\
      CoPRS~\cite{lu2025coprs}            & Qwen2.5-VL-3B   & 61.3 & \textbf{60.6} & 57.8 & \textbf{52.7} \\
      \rowcolor{gray!15}
      \textbf{DGSeg} & Qwen2.5-VL-3B & \textbf{66.0} & 58.3 & \textbf{60.0} & 50.3 \\
      \midrule
      LISA~\cite{lai2024lisa}             & LLaVA1.5-7B     & 53.6 & 52.3 & 48.7 & 48.8 \\
      SAM4MLLM~\cite{chen2024sam4mllm}    & Qwen-VL-7B      & 46.1 & 48.1 & --   & --   \\
      RSVP~\cite{lu2025rsvp}              & Qwen2-VL-7B     & 58.6 & 48.5 & 56.6 & 51.6 \\
      Seg-Zero~\cite{liu2025seg}          & Qwen2.5-VL-7B   & 62.6 & 62.0 & 57.5 & 52.0 \\
      SAM-R1~\cite{huang2025sam}          & Qwen2.5-VL-7B   & 54.0 & 55.8 & 60.2 & 54.3 \\
      Seg-R1~\cite{you2025seg}            & Qwen2.5-VL-7B   & 58.6 & 41.2 & 56.7 & 53.7 \\
      RISE~\cite{zhou2026reasoning}              & Qwen2.5-VL-7B   & 62.0 & 55.3 & 58.7 & 52.5 \\
      PIXELTHINK~\cite{wang2025pixelthink}& Qwen2.5-VL-7B   & 63.8 & 62.7 & 60.2 & 55.8 \\
      CoPRS~\cite{lu2025coprs}            & Qwen2.5-VL-7B   & 65.2 & 64.5 & 59.8 & 55.1 \\
      SAM-Veteran~\cite{dusam}            & Qwen2.5-VL-7B   & \underline{68.2} & \textbf{67.3} & \underline{62.6} & 56.1 \\
      SAM3 Agent~\cite{carion2025sam}     & Qwen2.5-VL-7B   & 65.4 & 50.5 & \underline{62.6}  & \underline{56.2} \\
      \rowcolor{gray!15}
      \textbf{DGSeg} & Qwen2.5-VL-7B & \textbf{69.6} & \underline{65.5} & \textbf{67.3} & \textbf{63.6} \\
      \bottomrule
    \end{tabular}%
  }

  \label{tab1:reasonseg_main}
\end{table}

\section{Experiment}
\subsection{Experimental Setup}

\textbf{Implementation Details.} We use Qwen2.5-VL~\cite{bai2025qwen2} as the reasoning model and SAM3~\cite{carion2025sam} as the segmentation backbone. In Stage 1, we finetune the MLLM with GRPO~\cite{shao2024deepseekmath} using a per-GPU batch size of 8 and sample each training instance 8 times. Furthermore, we finetune the MLLM using LoRA~\cite{hu2022lora}, and the number of trainable parameters is substantially fewer than in prior reasoning segmentation approaches~\cite{liu2025seg, zhu2025lens}. In Stage 2, we freeze the MLLM and SAM3 and train only the dynamic gating module for 3 epochs with a batch size of 1, using a learning rate of $1\times10^{-4}$ and a weight decay of $1\times10^{-4}$. All experiments are conducted on two NVIDIA H100 GPUs with 80GB memory each. More details can be found in the supplementary material.

\noindent\textbf{Dataset and Evaluation Metrics.} We train both stages using 9,000 instances sampled from RefCOCOg~\cite{yu2016modeling}. Following \cite{lai2024lisa, liu2025seg}, we evaluate on ReasonSeg~\cite{lai2024lisa}, reporting both gIoU and cIoU, as well as on RefCOCO~\cite{kazemzadeh2014referitgame}, RefCOCO+~\cite{yu2016modeling}, and RefCOCOg~\cite{yu2016modeling},  reporting cIoU as the evaluation metric. All experiments on the ReasonSeg dataset follow the zero-shot setting as there are no reasoning segmentation instances used during training. Since cIoU is computed as the ratio between the summed intersection and summed union over all images, it is biased towards large images~\cite{lai2024lisa, shen2025reasoning}, so we primarily focus on gIoU for ReasonSeg. All results in experiments are in \%.

\begin{table}[t]
  \centering
  \small
  \caption{Performance comparison on RefCOCO, RefCOCO+ and RefCOCOg benchmarks.}
  \setlength{\tabcolsep}{2.6pt}
  \renewcommand{\arraystretch}{1.03}

  \makebox[0.92\linewidth][c]{%
    \begin{tabular}{l >{\centering\arraybackslash}p{2.5cm} c c c c}
      \toprule
      \multirow{2}{*}{\textbf{Method}} &
      \multirow{2}{*}{\textbf{MLLM}} &
      \textbf{RefCOCO} &
      \textbf{RefCOCO+} &
      \textbf{RefCOCOg} &
      \multirow{2}{*}{\textbf{Avg.}} \\
      & & \textbf{testA} & \textbf{testA} & \textbf{test} & \\
      \midrule
      LAVT~\cite{yang2022lavt}              & --            & 75.8 & 68.4 & 62.1 & 68.8 \\
      ReLA~\cite{liu2023gres}              & --            & 76.5 & 71.0 & 66.0 & 71.2 \\
      \midrule
      Seg-Zero~\cite{liu2025seg}           & Qwen2.5-VL-3B  & \textbf{79.3} & \underline{73.7} & 71.5 & \underline{74.8} \\
      Seg-R1~\cite{you2025seg}             & Qwen2.5-VL-3B  & 76.0 & 66.8 & 67.9 & 70.2 \\
      RISE~\cite{zhou2026reasoning}                & Qwen2.5-VL-3B  & 76.5 & 72.1 & 68.9 & 72.5 \\
      PIXELTHINK~\cite{wang2025pixelthink} & Qwen2.5-VL-3B  & 78.7 & 72.9 & \underline{72.2} & 74.6 \\
      \rowcolor{gray!15}
      \textbf{DGSeg}                         & Qwen2.5-VL-3B  & \underline{78.9} & \textbf{74.5} & \textbf{72.3} & \textbf{75.2} \\
      \midrule
      LISA~\cite{lai2024lisa}           & LLaVA-7B       & 76.5 & 67.4 & 68.5 & 70.8 \\
      PixelLM~\cite{ren2024pixellm}        & LLaVA-7B       & 76.5 & 71.7 & 70.5 & 72.9 \\
      Seg-Zero~\cite{liu2025seg}           & Qwen2.5-VL-7B  & \underline{80.3} & 76.2 & 72.6 & \underline{76.4} \\
      Seg-R1~\cite{you2025seg}             & Qwen2.5-VL-7B  & 78.7 & 70.9 & 71.4 & 73.7 \\
      SAM-R1~\cite{huang2025sam}           & Qwen2.5-VL-7B  & 79.2 & 74.7 & 73.1 & 75.7 \\
      RISE~\cite{zhou2026reasoning}               & Qwen2.5-VL-7B  & 79.7 & \textbf{77.7} & \underline{73.4} & \textbf{76.9} \\
      PIXELTHINK~\cite{wang2025pixelthink} & Qwen2.5-VL-7B  & 79.3 & 74.8 & \textbf{73.9} & 76.0 \\
      SAM-Veteran~\cite{dusam}             & Qwen2.5-VL-7B  & \textbf{80.8} & \underline{76.6} & \underline{73.4} & \textbf{76.9} \\
      SAM3 Agent~\cite{carion2025sam}      & Qwen2.5-VL-7B  & 58.4 & 52.2 & 55.1 & 55.2 \\
      \rowcolor{gray!15}
      \textbf{DGSeg}                         & Qwen2.5-VL-7B  & \underline{80.3} & 76.4 & \textbf{73.9} & \textbf{76.9} \\
      \bottomrule
    \end{tabular}%
  }

  \label{tab2:refcoco_main}
\end{table}

\subsection{Main Results}
\noindent\textbf{Results on Reasoning Segmentation Benchmarks.} \Cref{tab1:reasonseg_main} reports the zero-shot performance on ReasonSeg, where DGSeg demonstrates remarkable effectiveness.  With the 3B-scale model, DGSeg achieves \textbf{66.0}/\textbf{60.0}\% gIoU on the validation and test splits. Compared with recent strong baselines built on the same MLLM backbone, DGSeg improves over Seg-Zero~\cite{liu2025seg} by 3.4\% and 3.9\% gIoU on the validation and test splits, respectively, and surpasses CoPRS~\cite{lu2025coprs} by 4.7\% and 2.2\% gIoU. The advantages of our design are further amplified when scaled to a 7B backbone, where DGSeg reaches  \textbf{69.6}/\textbf{67.3}\% gIoU,  surpassing SAM-Veteran~\cite{dusam} by 1.4\% and 4.7\%  on the validation and test splits, respectively.  

\noindent\textbf{Results on Referring Segmentation Benchmarks.} \Cref{tab2:refcoco_main} reports the performance on the RefCOCO, RefCOCO+ and RefCOCOg datasets. DGSeg achieves strong results across all splits, reaching a leading average score of 75.2\% with the 3B model and 76.9\% with the 7B model. Compared with recent baselines built on the same MLLM backbone, DGSeg improves over Seg-Zero~\cite{liu2025seg} on average. These results demonstrate that DGSeg ensures the final segmentation aligns with the reasoning intent by leveraging semantic-spatial cues and adaptive fusion on dual-branch predictions.

\begin{table}[t]
    \centering
    \caption{Ablation on fusion strategies within the proposed pipeline.}
    \label{tab:fusion_strategy}
    \begin{tabular}{lcc}
        \toprule
        \textbf{Variant} & \textbf{Val} & \textbf{Test} \\
        \midrule
        Oracle fusion & 67.9 & 62.8 \\
        \midrule
        Baseline (no fusion) & 60.8 & 55.7 \\
        \midrule
        \multicolumn{3}{@{}l}{\footnotesize\textit{Non-learnable fusion}} \\
        Ours + average fusion & 63.8 & 58.2 \\
        Ours + confidence-based weighting & 64.1 & 57.2 \\
        \midrule
                \rowcolor{gray!15}
        \multicolumn{3}{@{}l}{\footnotesize\textit{Learnable fusion}} \\
            \rowcolor{gray!15}
        Ours + learned scalar & 64.0 & 57.8 \\
        \rowcolor{gray!15}
        Ours + MoE & 65.6 & \textbf{60.6} \\
        \rowcolor{gray!15}
        \textbf{Ours + dynamic gating} & \textbf{66.0} & 60.0 \\
        \bottomrule
    \end{tabular}
\end{table}

\subsection{Ablation Study}

We conduct ablations to analyze the effectiveness of our proposed pipeline and DGSeg design. All experiments are evaluated on the model based on Qwen2.5-VL-3B~\cite{bai2025qwen2}, the benchmark is  ReasonSeg, and the metric is gIoU. More ablation experimental details and results can be found in the supplementary material.

\noindent\textbf{Ablation on the Proposed Pipeline.} We first evaluate the effectiveness of the proposed pipeline by comparing different fusion strategies in \cref{tab:fusion_strategy}. Oracle fusion uses the soft target weight $W^{\ast}$ as an upper bound, while the baseline directly encodes semantic and spatial cues without fusion. To further analyze the role of fusion, we compare both non-learnable strategies, including average fusion and confidence-based weighting, and learnable strategies, including learned scalar, MoE, and dynamic gating.

Results show that learnable fusion strategies generally achieve stronger performance than non-learnable ones, demonstrating the effectiveness of learning the fusion process from the training data. By learning the relationship between pixel-level branch features and segmentation quality, these strategies can better assess the relative reliability of different cues and adapt the fusion weights accordingly. Notably, MoE achieves comparable performance to dynamic gating, suggesting that the proposed pipeline can flexibly accommodate advanced learnable fusion strategies. The integration of more advanced fusion methods into our pipeline can be explored for future work.

\noindent\textbf{Analysis of Reward Design.} We next analyze the reward design used in the RL finetuning stage and summarize the results in \cref{tab:ablation_reward}. Here, ``label reward'' uses category labels as the category label supervision, while the semantic reward in DGSeg is based on the segmentation model preference. The results indicate that DGSeg benefits from jointly optimizing the output format and the quality of the generated cues. Specifically, the preference optimization based on SAM3 yields better results when applied to semantic cues rather than directly to category labels. This suggests that, to tightly couple the reasoning model with the segmentation model, it is more important to generate semantic cues that are better aligned with the prompt space of the segmentation model.

\noindent\textbf{Analysis of Dual-Branch Segmentation Design.} We then analyze the dual-branch segmentation design in \cref{tab:ablation_fusion}. The baseline with no fusion follows the standard prompting scheme of SAM3 by encoding semantic and spatial cues jointly in a single forward pass. To examine the contribution of each cue type, we further report semantic-only and spatial-only variants, which perform segmentation using only the semantic or spatial branch, respectively. Results indicate that joint encoding can allow noise from any cue to contaminate the prediction,  leading to performance that can even be worse than using the spatial cue alone. In contrast, DGSeg evaluates the pixel-level features from dual branches and suppresses noisy regions, yielding the best overall performance. 

Additionally, we analyze the computational overhead introduced by the dual-branch architecture. As shown in \cref{tab:efficiency}, our proposed components incur only a 0.3\% increase in FLOPs and a 2.0\% drop in FPS, while effectively boosting the overall reasoning segmentation performance.

\begin{table*}[t]
  \centering
  \small
  \setlength{\tabcolsep}{3pt}
  \renewcommand{\arraystretch}{1.05}

  \begin{minipage}[t]{0.49\linewidth}
    \centering
    \caption{Ablation on the reward design.}
    \makebox[\linewidth][c]{%
      \begin{tabular}{l cc}
        \toprule
        \textbf{Variant} & \textbf{Val} & \textbf{Test} \\
        \midrule
        w/o Stage 1                   & 36.0 & 33.7 \\
        \midrule
        w/o format reward                    & 64.6 & 59.6 \\
        w/o semantic reward                  & 62.1 & 58.5\\
        w/o spatial reward                   & 64.1 & 59.4 \\
        $R_{\text{semantic}}$ = label reward & 64.7 & 59.1 \\
        \midrule
        \rowcolor{gray!15}
        \textbf{Full reward (ours)}          & \textbf{66.0} & \textbf{60.0} \\
        \bottomrule
      \end{tabular}%
    }
    \label{tab:ablation_reward}
  \end{minipage}
  \hfill
  \begin{minipage}[t]{0.49\linewidth}
    \centering
    \caption{Ablation on the dual-branch segmentation strategy.}
    \makebox[\linewidth][c]{%
      \begin{tabular}{p{4.1cm} cc}
        \toprule
        \textbf{Variant} & \textbf{Val} & \textbf{Test} \\
        \midrule
        Baseline (no fusion)                 & 60.8 & 55.7 \\
        \midrule
        Semantic--only                                        & 55.3 & 52.8 \\
        Spatial--only                                        & 63.4 & 54.9 \\
        \midrule
        \rowcolor{gray!15}
        \makecell[l]{\textbf{Learnable fusion (ours)}\\[-1pt]\footnotesize{\textbf{(dynamic gate)}}}
            & \textbf{66.0} & \textbf{60.0} \\
        \bottomrule
      \end{tabular}%
    }
    \label{tab:ablation_fusion}
  \end{minipage}
\end{table*}

\begin{figure}[t]
  \centering
  \begin{subfigure}[t]{0.48\linewidth}
    \centering
    \includegraphics[width=\linewidth]{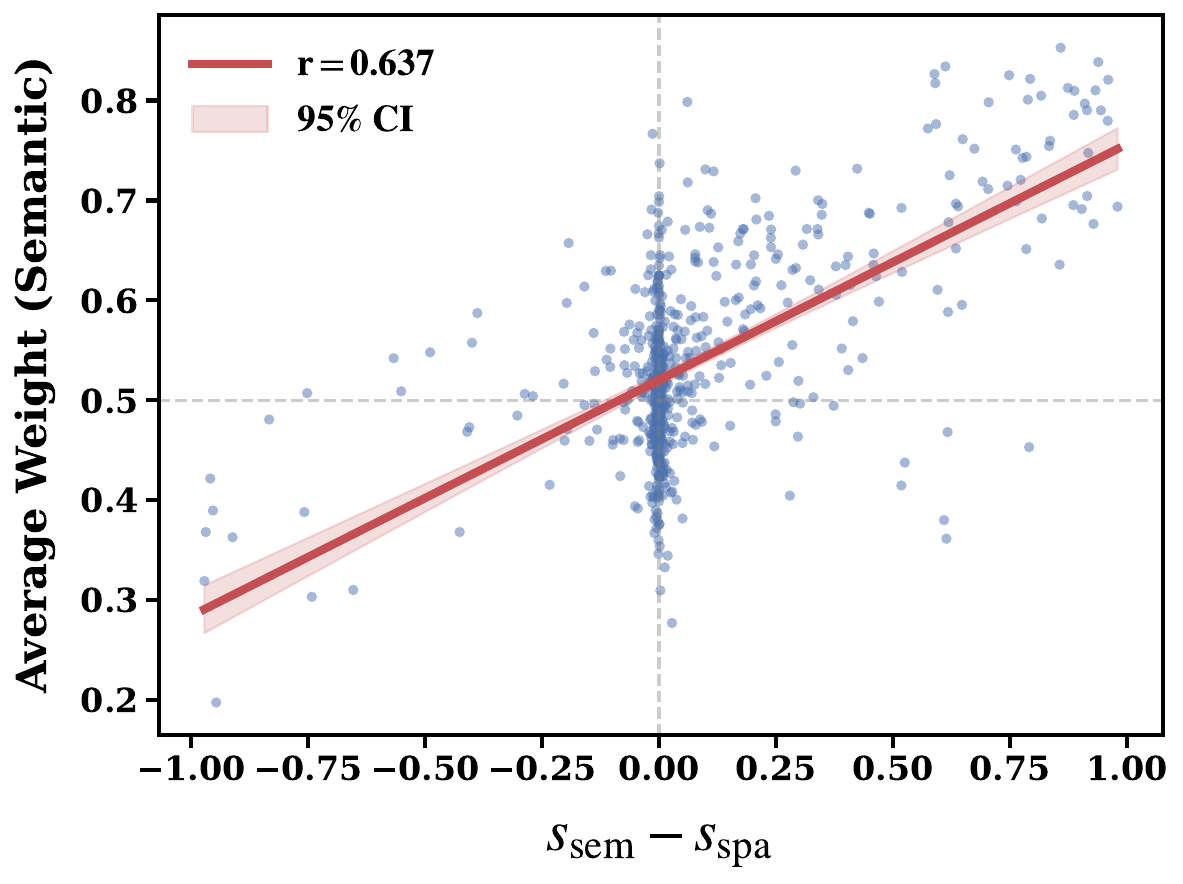}
    \caption{The relationship between IoU difference of the semantic and spatial branches and the average fusion weights of the semantic branch. ``CI'' means Confidence Interval.}
    \label{fig:fig3a}
  \end{subfigure}\hfill
  \begin{subfigure}[t]{0.49\linewidth}
    \centering
    \includegraphics[width=\linewidth]{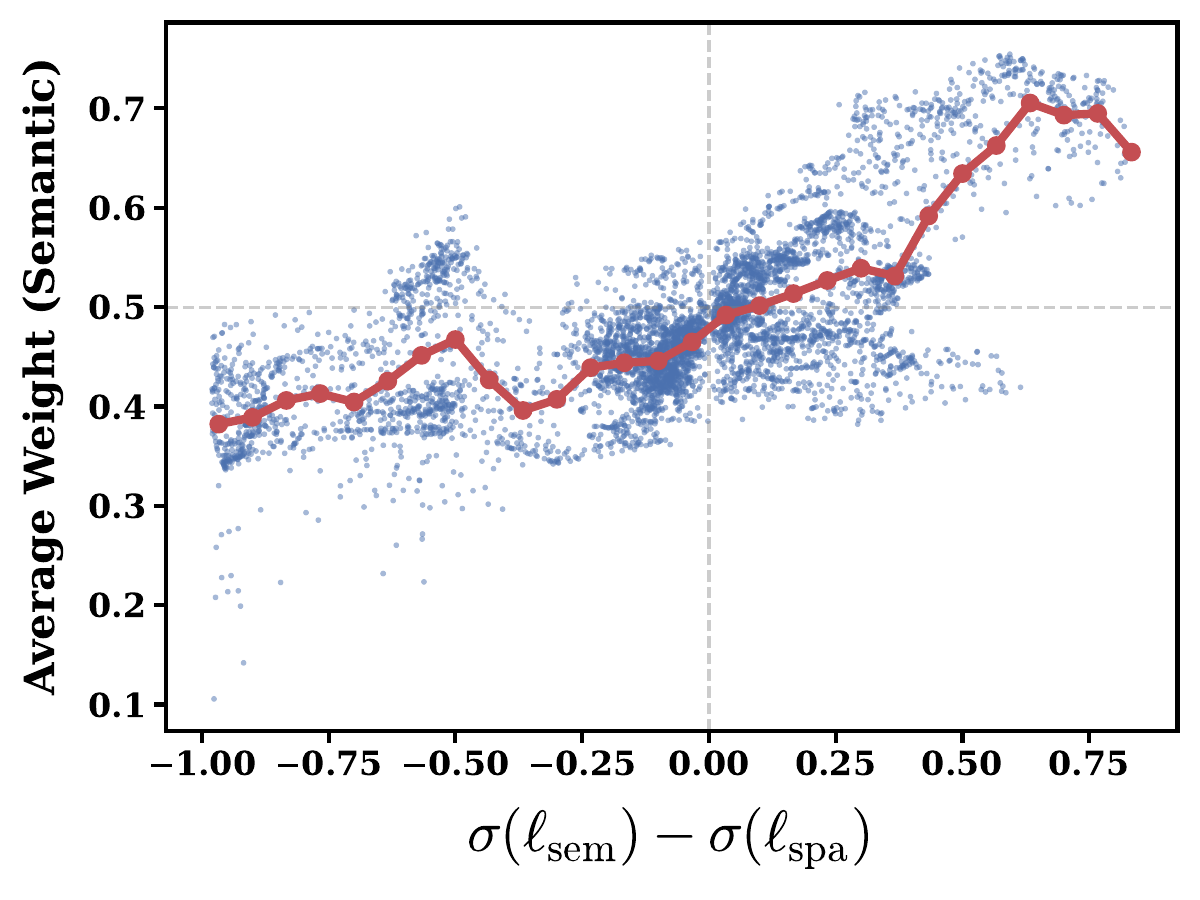}
    \caption{The relationship between sigmoid logit difference of the semantic and spatial branches and the average fusion weights of the semantic branch at the pixel level. }
    \label{fig:fig3b}
  \end{subfigure}
  \caption{Visualization of the analysis on the dynamic gating module.}
  \label{fig:fig3}
\end{figure}

\begin{figure}
  \centering
  \includegraphics[width=\textwidth]{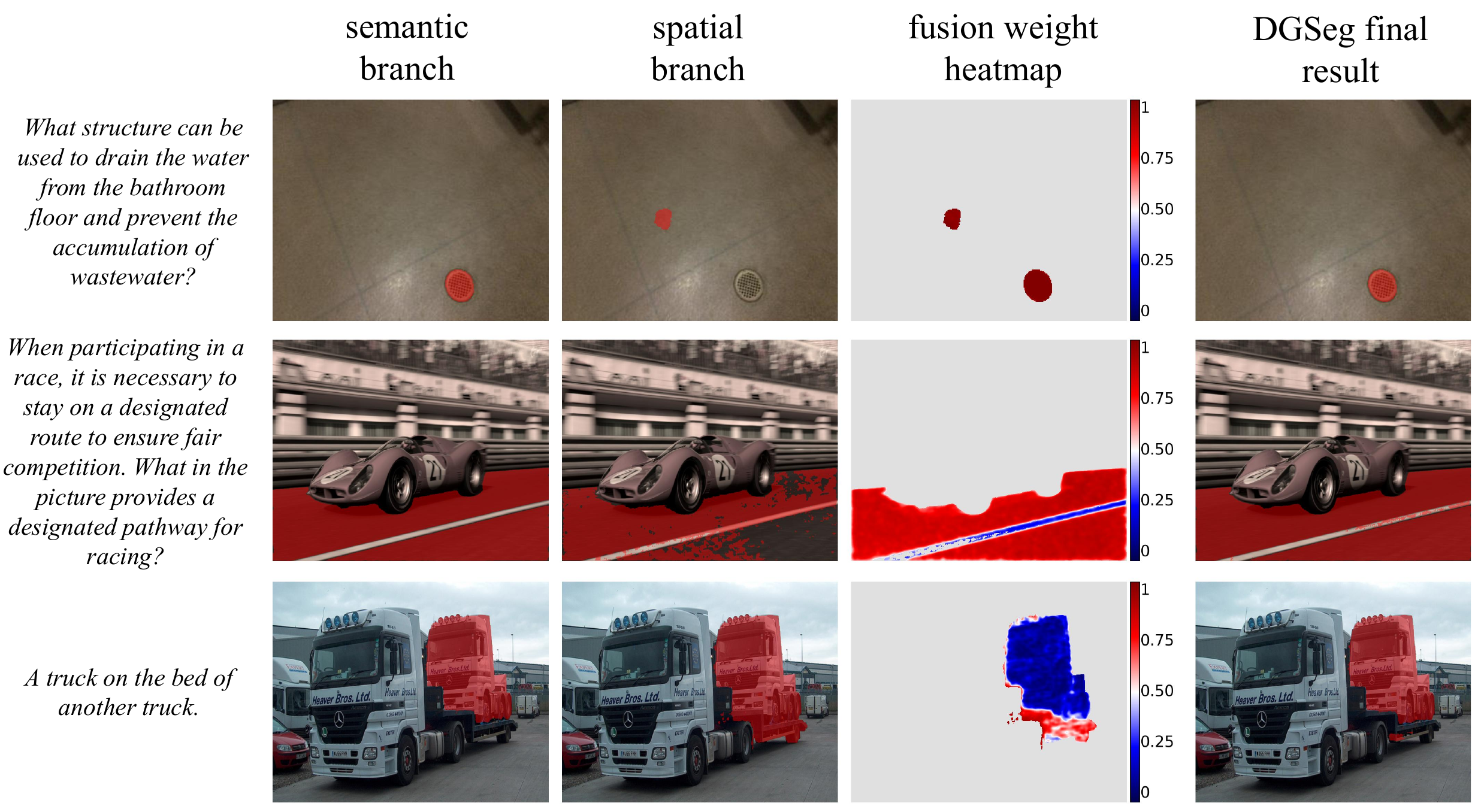}
  \caption{\textbf{Visualization of gating results in DGSeg.} For each example, we show the semantic and spatial branch predictions, final results after dynamic gating, and the foreground fusion weights of the semantic branch. Red favors the \textcolor{red}{semantic branch}, and blue favors the \textcolor{blue}{spatial branch}. We set all background regions to gray for better visualization.}
  \label{fig4}
\end{figure}

\begin{table}[t]
  \centering
  \small
  \caption{Analysis of computational efficiency. We report peak GPU memory usage, floating-point operations (FLOPs), and frames per second (FPS) for the no-fusion baseline and DGSeg, measured on a single NVIDIA H100 GPU.}
  \setlength{\tabcolsep}{8pt}
  \renewcommand{\arraystretch}{1.03}

  \makebox[0.95\linewidth][c]{%
    \begin{tabular}{l c c c}
      \toprule
      \textbf{Method}  & Peak Memory (GB) $\downarrow$ & FLOPs (GFLOPs) $\downarrow$ & FPS $\uparrow$  \\
      \midrule
      Baseline (no fusion)   & 11.75 & 7375.7 & 0.280\\
\rowcolor{gray!15}
      \textbf{DGSeg}  & 12.24 & 7395.1  & 0.275\\
    \rowcolor{gray!15}
      &  \textbf{{\scriptsize (+4.2\%)}} & \textbf{{\scriptsize (+0.3\%)}} & \textbf{{\scriptsize (-2.0\%)}}\\
      \bottomrule
    \end{tabular}%
  }
  \label{tab:efficiency}
\end{table}

\begin{figure}
  \centering
  \includegraphics[width=\textwidth]{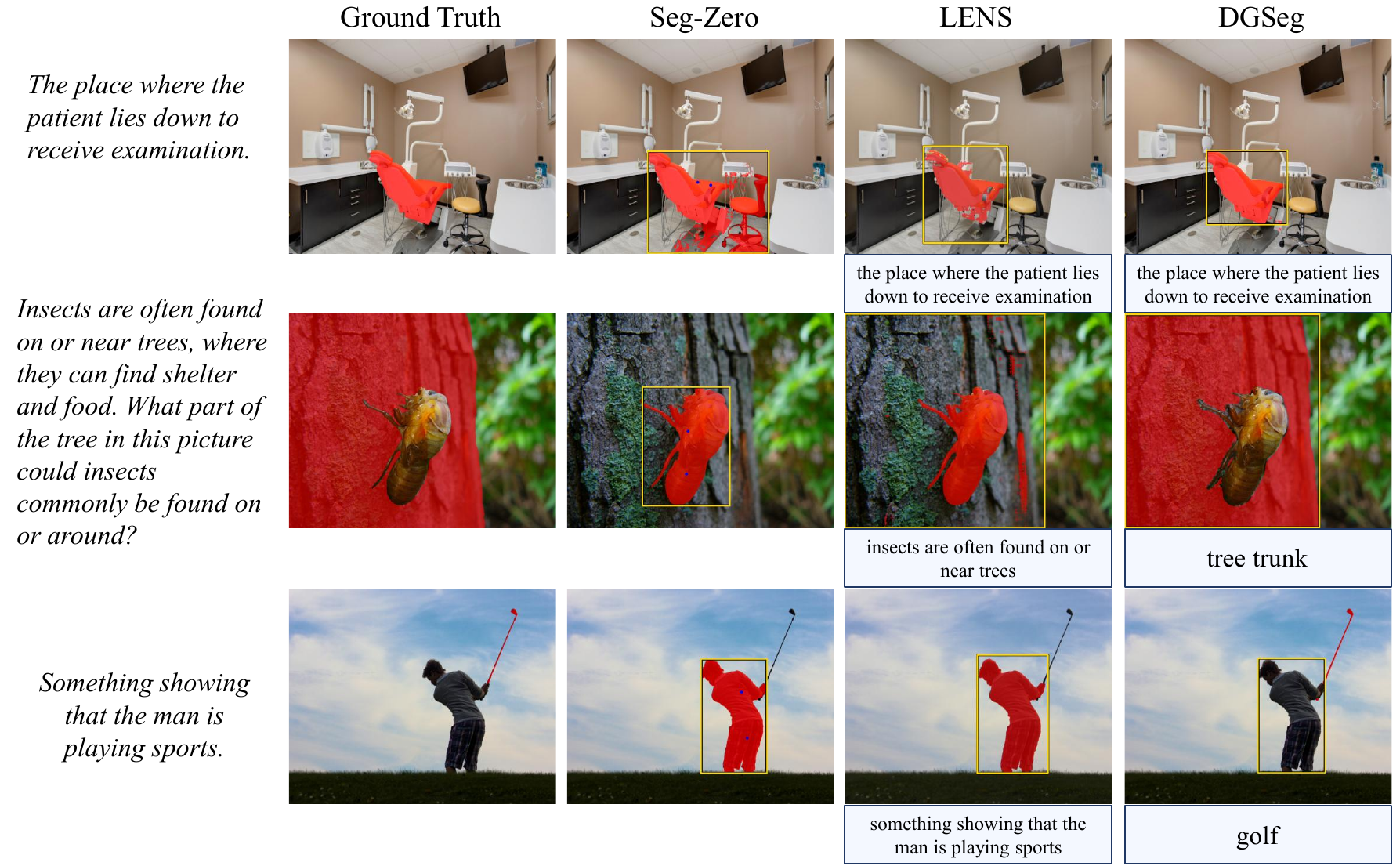}
  \caption{\textbf{Qualitative comparison on ReasonSeg.} We compare strong baselines including Seg-Zero~\cite{liu2025seg} and LENS~\cite{zhu2025lens}. For each example, the left shows the query and the right visualizes the predictions and the generated target cues. DGSeg produces semantic-spatial cues and performs adaptive fusion, leading to better performance even when some cues are ambiguous or noisy.}
  \label{fig5}
\end{figure}

\noindent\textbf{Correlation Analysis of Fusion Weights.}  We further examine whether the proposed learnable fusion module in DGSeg effectively assesses segmentation features and filters unreliable regions from both global and local perspectives.

For global correlation, we compute the IoU of the semantic and spatial branch predictions with the ground truth mask, and use their IoU difference $s_{\text{sem}} - s_{\text{spa}}$ as a proxy for relative branch quality, where a larger gap indicates that the predictions of the semantic branch are closer to ground truth than those of the spatial branch.  We also compute the average fusion weight over foreground regions where fusion is more meaningful. As shown in \cref{fig:fig3a}, the IoU difference exhibits a strong positive correlation ($r = 0.637$) with the predicted fusion weight. This quantitative relationship indicates that by learning to analyze joint feature representations of the image and the cues, the module can effectively identify the superior branch and assign appropriate fusion weights.

For local correlation, we analyze foreground pixels by randomly sampling those where the predictions from the two branches are inconsistent (i.e., exhibiting a sign mismatch in logits), and we correlate the fusion weights with the sigmoid logit difference. \Cref{fig:fig3b} shows that when semantic evidence is stronger (positive logit difference), the predicted weight tends to favor the semantic branch, and vice versa. This confirms that by learning to evaluate the features from the segmentation branches, the dynamic gating module evaluates the distribution of local high response regions and boundary structures, enabling it to perform effective local denoising.

\subsection{Qualitative Analysis}

\noindent\textbf{Visualization of Gating Results.} \Cref{fig4} presents qualitative comparisons between the two branches and the final results. As shown, relying solely on either semantic or spatial cues often results in ambiguous or even erroneous target references. Consequently, the segmentation model is prone to producing incorrect areas within the mask. In contrast, our proposed dynamic gating module effectively identifies unreliable regions and adaptively adjusts the weights of the two branches. For example, in the bottom case, it detects the erroneous segmentation of the larger truck from the spatial branch and increases the weight of the semantic branch in that area.

\noindent\textbf{Qualitative Comparisons.} \Cref{fig5} presents visual comparisons between DGSeg-3B and representative baselines including Seg-Zero~\cite{liu2025seg} and LENS~\cite{zhu2025lens} on challenging cases of ReasonSeg~\cite{lai2024lisa}. Seg-Zero provides purely spatial prompts (a bounding box and two points) to guide the downstream segmentation model, which can lead to ambiguous referents and irrelevant regions. For example, the predicted mask in the first row includes unrelated objects due to the oversized bounding box. While LENS also shares a strategy of combining semantic and spatial information and has been finetuned on the ReasonSeg training split, it lacks a mechanism to assess target cues and is highly susceptible to erroneous information  (evidenced by the failure case in the second row).  In contrast,  DGSeg adaptively suppresses noisy branches and remains robust even when some target cues are ambiguous or noisy (e.g., the spatial cue fails to cover the golf).

\section{Conclusion}
In this paper, we propose a novel reasoning segmentation pipeline based on complementary cue processing and learnable adaptive fusion, aiming to reduce the impact of ambiguous or erroneous cues in MLLM-guided segmentation. We instantiate this pipeline as DGSeg, which generates semantic and spatial cues from the MLLM, processes them with separate segmentation branches, and aggregates their predictions using a lightweight dynamic gating module. Extensive experiments on reasoning and referring segmentation benchmarks demonstrate the effectiveness of DGSeg, while ablation studies further validate the benefit of the proposed pipeline and its learnable fusion process. Our analysis also shows that the pipeline can accommodate different advanced fusion strategies, suggesting a promising direction for future exploration of more effective fusion designs in reasoning segmentation.

\section*{Acknowledgements}
This research is supported by the National Key Research and Development Program of China under Grant No. 2025YFE0216600 and Key Science \& Technology Project of Anhui Province under Grant No. 202523j08050027.

\bibliographystyle{splncs04}
\bibliography{main}

\begin{thebibliography}{10}
\providecommand{\url}[1]{\texttt{#1}}
\providecommand{\urlprefix}{URL }
\providecommand{\doi}[1]{https://doi.org/#1}

\bibitem{bai2023qwen1}
Bai, J., Bai, S., Chu, Y., Cui, Z., Dang, K., Deng, X., Fan, Y., Ge, W., Han, Y., Huang, F., et~al.: Qwen technical report. arXiv preprint arXiv:2309.16609  (2023)

\bibitem{bai2023qwen}
Bai, J., Bai, S., Yang, S., Wang, S., Tan, S., Wang, P., Lin, J., Zhou, C., Zhou, J.: Qwen-vl: A versatile vision-language model for understanding, localization, text reading, and beyond. arXiv preprint arXiv:2308.12966  (2023)

\bibitem{bai2025qwen2}
Bai, S., Chen, K., Liu, X., Wang, J., Ge, W., Song, S., Dang, K., Wang, P., Wang, S., Tang, J., et~al.: Qwen2.5-vl technical report. arXiv preprint arXiv:2502.13923  (2025)

\bibitem{bao2024cores}
Bao, X., Sun, S., Ma, S., Zheng, K., Guo, Y., Zhao, G., Zheng, Y., Wang, X.: Cores: Orchestrating the dance of reasoning and segmentation. In: ECCV (2024)

\bibitem{carion2025sam}
Carion, N., Gustafson, L., Hu, Y.T., Debnath, S., Hu, R., Suris, D., Ryali, C., Alwala, K.V., Khedr, H., Huang, A., et~al.: Sam 3: Segment anything with concepts. arXiv preprint arXiv:2511.16719  (2025)

\bibitem{carion2020end}
Carion, N., Massa, F., Synnaeve, G., Usunier, N., Kirillov, A., Zagoruyko, S.: End-to-end object detection with transformers. In: ECCV (2020)

\bibitem{chen2024spatialvlm}
Chen, B., Xu, Z., Kirmani, S., Ichter, B., Sadigh, D., Guibas, L., Xia, F.: Spatialvlm: Endowing vision-language models with spatial reasoning capabilities. In: CVPR (2024)

\bibitem{chen2024sam4mllm}
Chen, Y.C., Li, W.H., Sun, C., Wang, Y.C.F., Chen, C.S.: Sam4mllm: Enhance multi-modal large language model for referring expression segmentation. In: ECCV (2024)

\bibitem{chen2024expanding}
Chen, Z., Wang, W., Cao, Y., Liu, Y., Gao, Z., Cui, E., Zhu, J., Ye, S., Tian, H., Liu, Z., et~al.: Expanding performance boundaries of open-source multimodal models with model, data, and test-time scaling. arXiv preprint arXiv:2412.05271  (2024)

\bibitem{chen2024internvl}
Chen, Z., Wu, J., Wang, W., Su, W., Chen, G., Xing, S., Zhong, M., Zhang, Q., Zhu, X., Lu, L., et~al.: Internvl: Scaling up vision foundation models and aligning for generic visual-linguistic tasks. In: CVPR (2024)

\bibitem{christiano2017deep}
Christiano, P.F., Leike, J., Brown, T.B., Martic, M., Legg, S., Amodei, D.: Deep reinforcement learning from human preferences. In: NeurIPS (2017)

\bibitem{collins2024forcesight}
Collins, J.A., Houff, C., Tan, Y.L., Kemp, C.C.: Forcesight: Text-guided mobile manipulation with visual-force goals. In: ICRA (2024)

\bibitem{dai2023instructblip}
Dai, W., Li, J., Li, D., Tiong, A.M.H., Zhao, J., Wang, W., Li, B., Fung, P., Hoi, S.: Instructblip: Towards general-purpose vision-language models with instruction tuning. In: NeurIPS (2023)

\bibitem{deng2009imagenet}
Deng, J., Dong, W., Socher, R., Li, L.J., Li, K., Fei-Fei, L.: Imagenet: A large-scale hierarchical image database. In: CVPR (2009)

\bibitem{dusam}
Du, T., Li, H., Fan, Z., Zhang, J., Pan, P., Zhang, Y.: Sam-veteran: An mllm-based human-like sam agent for reasoning segmentation. In: ICLR (2026)

\bibitem{fan2023stable}
Fan, Q., Tao, X., Ke, L., Ye, M., Zhang, D., Wan, P., Tai, Y.W., Tang, C.K.: Stable segment anything model. In: ICLR (2025)

\bibitem{he2025analyzing}
He, H., Li, G., Geng, Z., Xu, J., Peng, Y.: Analyzing and boosting the power of fine-grained visual recognition for multi-modal large language models. In: ICLR (2025)

\bibitem{he2025rsagent}
He, X., Zhang, Y., Gao, S., Li, W., Hong, L., Chen, M., Jiang, K., Fu, J., Zhang, W.: Rsagent: Learning to reason and act for text-guided segmentation via multi-turn tool invocations. arXiv preprint arXiv:2512.24023  (2025)

\bibitem{hu2022lora}
Hu, E.J., Shen, Y., Wallis, P., Allen-Zhu, Z., Li, Y., Wang, S., Wang, L., Chen, W.: Lora: Low-rank adaptation of large language models. In: ICLR (2022)

\bibitem{hu2024visual}
Hu, Y., Stretcu, O., Lu, C.T., Viswanathan, K., Hata, K., Luo, E., Krishna, R., Fuxman, A.: Visual program distillation: Distilling tools and programmatic reasoning into vision-language models. In: CVPR (2024)

\bibitem{huang2025sam}
Huang, J., Xu, Z., Zhou, J., Liu, T., Xiao, Y., Ou, M., Ji, B., Li, X., Yuan, K.: Sam-r1: Leveraging sam for reward feedback in multimodal segmentation via reinforcement learning. In: NeurIPS (2025)

\bibitem{jia2021scaling}
Jia, C., Yang, Y., Xia, Y., Chen, Y.T., Parekh, Z., Pham, H., V.Le, Q., Sung, Y., Li, Z., Duerig, T.: Scaling up visual and vision-language representation learning with noisy text supervision. In: ICML (2021)

\bibitem{kang2025your}
Kang, S., Kim, J., Kim, J., Hwang, S.J.: Your large vision-language model only needs a few attention heads for visual grounding. In: CVPR (2025)

\bibitem{kazemzadeh2014referitgame}
Kazemzadeh, S., Ordonez, V., Matten, M., Berg, T.L.: Referitgame: Referring to objects in photographs of natural scenes. In: EMNLP (2014)

\bibitem{kirillov2023segment}
Kirillov, A., Mintun, E., Ravi, N., Mao, H., Rolland, C., Gustafson, L., Xiao, T., Whitehead, S., Berg, A.C., Lo, W.Y., et~al.: Segment anything. In: ICCV (2023)

\bibitem{lai2024lisa}
Lai, X., Tian, Z., Chen, Y., Li, Y., Yuan, Y., Liu, S., Jia, J.: Lisa: Reasoning segmentation via large language model. In: CVPR (2024)

\bibitem{li2021align}
Li, J., Selvaraju, R.R., Gotmare, A.D., Joty, S., Xiong, C., Hoi, S.C.H.: Align before fuse: Vision and language representation learning with momentum distillation. In: NeurIPS (2021)

\bibitem{liang2023open}
Liang, F., Wu, B., Dai, X., Li, K., Zhao, Y., Zhang, H., Zhang, P., Vajda, P., Marculescu, D.: Open-vocabulary semantic segmentation with mask-adapted clip. In: CVPR (2023)

\bibitem{lin2025samrefiner}
Lin, Y., Li, H., Shao, W., Yang, Z., Zhao, J., He, X., Luo, P., Zhang, K.: Samrefiner: Taming segment anything model for universal mask refinement. In: ICLR (2025)

\bibitem{liu2023gres}
Liu, C., Ding, H., Jiang, X.: Gres: Generalized referring expression segmentation. In: CVPR (2023)

\bibitem{liu2023visual}
Liu, H., Li, C., Wu, Q., Lee, Y.J.: Visual instruction tuning. In: NeurIPS (2023)

\bibitem{liu2025segmentation}
Liu, J., Chen, L.: Segmentation as a plug-and-play capability for frozen multimodal llms. arXiv preprint arXiv:2510.16785  (2025)

\bibitem{liu2025seg}
Liu, Y., Peng, B., Zhong, Z., Yue, Z., Lu, F., Yu, B., Jia, J.: Seg-zero: Reasoning-chain guided segmentation via cognitive reinforcement. arXiv preprint arXiv:2503.06520  (2025)

\bibitem{liu2025visionreasoner}
Liu, Y., Qu, T., Zhong, Z., Peng, B., Liu, S., Yu, B., Jia, J.: Visionreasoner: Unified reasoning-integrated visual perception via reinforcement learning. arXiv preprint arXiv:2505.12081  (2025)

\bibitem{lu2025rsvp}
Lu, Y., Cao, J., Wu, Y., Li, B., Tang, L., Ji, Y., Wu, C., Wu, J., Zhu, W.: Rsvp: Reasoning segmentation via visual prompting and multi-modal chain-of-thought. In: ACL (2025)

\bibitem{lu2025coprs}
Lu, Z., Li, L., Wang, J., Feng, Y., Chen, B., Chen, K., Wang, Y.: Coprs: Learning positional prior from chain-of-thought for reasoning segmentation. arXiv preprint arXiv:2510.11173  (2025)

\bibitem{man1982computational}
Marr, D.: Vision: A computational investigation into the human representation and processing of visual information. MIT press (2010)

\bibitem{qian2025reasoning}
Qian, R., Yin, X., Dou, D.: Reasoning to attend: Try to understand how< seg> token works. In: CVPR (2025)

\bibitem{radford2021learning}
Radford, A., Kim, J.W., Hallacy, C., Ramesh, A., Goh, G., Agarwal, S., Sastry, G., Askell, A., Mishkin, P., Clark, J., et~al.: Learning transferable visual models from natural language supervision. In: ICML (2021)

\bibitem{ravi2024sam}
Ravi, N., Gabeur, V., Hu, Y.T., Hu, R., Ryali, C., Ma, T., Khedr, H., R{\"a}dle, R., Rolland, C., Gustafson, L., et~al.: Sam 2: Segment anything in images and videos. In: ICLR (2025)

\bibitem{ren2024grounded}
Ren, T., Liu, S., Zeng, A., Lin, J., Li, K., Cao, H., Chen, J., Huang, X., Chen, Y., Yan, F., et~al.: Grounded sam: Assembling open-world models for diverse visual tasks. arXiv preprint arXiv:2401.14159  (2024)

\bibitem{ren2024pixellm}
Ren, Z., Huang, Z., Wei, Y., Zhao, Y., Fu, D., Feng, J., Jin, X.: Pixellm: Pixel reasoning with large multimodal model. In: CVPR (2024)

\bibitem{shao2024deepseekmath}
Shao, Z., Wang, P., Zhu, Q., Xu, R., Song, J., Bi, X., Zhang, H., Zhang, M., Li, Y., Wu, Y., et~al.: Deepseekmath: Pushing the limits of mathematical reasoning in open language models. arXiv preprint arXiv:2402.03300  (2024)

\bibitem{shen2025reasoning}
Shen, Y., Li, C., Xiong, F., Jeong, J.O., Wang, T., Latman, M., Unberath, M.: Reasoning segmentation for images and videos: A survey. arXiv preprint arXiv:2505.18816  (2025)

\bibitem{touvron2023llama}
Touvron, H., Lavril, T., Izacard, G., Martinet, X., Lachaux, M.A., Lacroix, T., Rozi{\`e}re, B., Goyal, N., Hambro, E., Azhar, F., et~al.: Llama: Open and efficient foundation language models. arXiv preprint arXiv:2302.13971  (2023)

\bibitem{vuong2024language}
Vuong, A.D., Vu, M.N., Huang, B., Nguyen, N., Le, H., Vo, T., Nguyen, A.: Language-driven grasp detection. In: CVPR (2024)

\bibitem{wang2024qwen2}
Wang, P., Bai, S., Tan, S., Wang, S., Fan, Z., Bai, J., Chen, K., Liu, X., Wang, J., Ge, W., et~al.: Qwen2-vl: Enhancing vision-language model's perception of the world at any resolution. arXiv preprint arXiv:2409.12191  (2024)

\bibitem{wang2025pixelthink}
Wang, S., Fang, G., Kong, L., Li, X., Xu, J., Yang, S., Li, Q., Zhu, J., Wang, X.: Pixelthink: Towards efficient chain-of-pixel reasoning. arXiv preprint arXiv:2505.23727  (2025)

\bibitem{wang2025cof}
Wang, Y., Gao, D., Li, B., Long, R., Yi, L., Cai, X., Yang, L., Zhang, J., Yu, S., Xuan, Q.: Cof: Coarse to fine-grained image understanding for multi-modal large language models. In: ICASSP (2025)

\bibitem{wei2022chain}
Wei, J., Wang, X., Schuurmans, D., Bosma, M., Xia, F., Chi, E.H., Le, Q.V., Zhou, D., et~al.: Chain-of-thought prompting elicits reasoning in large language models. In: NeurIPS (2022)

\bibitem{yan2024visa}
Yan, C., Wang, H., Yan, S., Jiang, X., Hu, Y., Kang, G., Xie, W., Gavves, E.: Visa: Reasoning video object segmentation via large language models. In: ECCV (2024)

\bibitem{yang2022lavt}
Yang, Z., Wang, J., Tang, Y., Chen, K., Zhao, H., Torr, P.H.: Lavt: Language-aware vision transformer for referring image segmentation. In: CVPR (2022)

\bibitem{yao2025argus}
Yao, Y., Li, L., Song, J., Chen, C., He, Z., Wang, Y., Wang, X., Gu, T., Li, J., Teng, Y., et~al.: Argus inspection: do multimodal large language models possess the eye of panoptes? In: ACMMM (2025)

\bibitem{you2025seg}
You, Z., Wu, Z.: Seg-r1: Segmentation can be surprisingly simple with reinforcement learning. arXiv preprint arXiv:2506.22624  (2025)

\bibitem{yu2016modeling}
Yu, L., Poirson, P., Yang, S., Berg, A.C., Berg, T.L.: Modeling context in referring expressions. In: ECCV (2016)

\bibitem{yuan2025sa2va}
Yuan, H., Li, X., Zhang, T., Sun, Y., Huang, Z., Xu, S., Ji, S., Tong, Y., Qi, L., Feng, J., Ming-Hsuan, Y.: Sa2va: Marrying sam2 with llava for dense grounded understanding of images and videos. arXiv preprint arXiv:2501.04001  (2025)

\bibitem{zhou2026reasoning}
Zhou, Q., Yang, L., Jia, Y., Gao, J., Ni, W., Wu, J., Wang, Q.: Reasoning via implicit self-supervised emergence for instruction segmentation. In: AAAI (2026)

\bibitem{zhu2025lens}
Zhu, L., Ouyang, B., Zhang, Y., Cheng, T., Hu, R., Shen, H., Ran, L., Chen, X., Yu, L., Liu, W., Xinggang, W.: Lens: Learning to segment anything with unified reinforced reasoning. In: AAAI (2026)

\bibitem{zhu2025segagent}
Zhu, M., Tian, Y., Chen, H., Zhou, C., Guo, Q., Liu, Y., Yang, M., Shen, C.: Segagent: Exploring pixel understanding capabilities in mllms by imitating human annotator trajectories. In: CVPR (2025)

\bibitem{zou2023segment}
Zou, X., Yang, J., Zhang, H., Li, F., Li, L., Wang, J., Wang, L., Gao, J., Lee, Y.J.: Segment everything everywhere all at once. In: NeurIPS (2023)

\end{thebibliography}
\end{document}